\documentclass{l4dc2027}
\usepackage{times}
\usepackage{graphicx}
\newtheorem{assumption}{Assumption}

\title[Bayesian $\kappa$-Explorer]{Bayesian $\kappa$-Explorer for Minimax Model Estimation in MDPs}
\author{%
 \Name{Xihe Gu} \Email{x9gu@ucsd.edu}\\
 \addr University of California, San Diego
 \AND
 \Name{Urbashi Mitra} \Email{ubli@usc.edu}\\
 \addr University of Southern California
 \AND
 \Name{Tara Javidi} \Email{tjavidi@ucsd.edu}\\
 \addr University of California, San Diego%
}

\begin{document}
\maketitle

\begin{abstract}
We study active exploration for minimax transition estimation in unknown Markov decision processes (MDPs).
Frequentist $\kappa$-Explorer~\citep{gu2026kappa} addresses this problem through a parameterized family of exploration objectives, but requires confidence-bound constructions and joint optimization over plausible transition kernels and occupancy measures, introducing a computational burden that limits its practical applicability.
We develop Bayesian $\kappa$-Explorer, which simplifies this procedure by placing Dirichlet priors on the transition distributions. Under the posterior mean estimator, the posterior expected squared $\ell_2$ error equals the total posterior variance, providing a closed-form exploration signal. The algorithm plans under the posterior mean model, eliminating confidence-set optimization.
Under the stated assumptions, this simpler planning procedure retains the $\widetilde{\mathcal O}(T^{-1/3})$ surrogate convergence rate of frequentist $\kappa$-Explorer.
We further bound the maximum posterior variance relative to the optimal asymptotic minimax allocation benchmark.
Experiments on Pendulum, Mountain Car, and Taxi demonstrate lower worst-case transition-estimation errors than the tested baselines.
\end{abstract}


\section{Introduction}
\label{sec:intro}
Accurately estimating the transition dynamics of an unknown Markov decision process (MDP) is a fundamental problem in reward-free reinforcement learning~\citep{araya2011active,tarbouriech2020active}. Under a limited interaction budget, observations must be allocated according to the intrinsic estimation complexity of each state--action pair: nearly deterministic transitions require fewer samples than transitions spread across many possible next states. To obtain uniformly accurate estimates, we therefore seek to minimize the largest transition-estimation error across state--action pairs~\citep{shekhar2020adaptive,shekhar2021adaptive,tarbouriech2020active,gu2026relative}.

Minimax estimation in MDPs is difficult for two distinct reasons. First, visitation frequencies are coupled through the dynamics, so increasing visits to one state--action pair affects the sampling allocation elsewhere. Prior work addresses coverage through state-visitation entropy~\citep{hazan2019provably}, trajectory entropy~\citep{tiapkin2023fast}, or more general concave objectives~\citep{gu2026relative}. However, broad coverage or matching a prescribed target does not necessarily yield the instance-dependent allocation required for minimax estimation. Second, the minimax objective is nonsmooth, preventing the direct application of methods designed for average estimation error or other smooth objectives~\citep{tarbouriech2020active}.

The frequentist $\kappa$-Explorer framework in~\cite{gu2026kappa} addresses these challenges through a family of exploration objectives, parameterized by a curvature exponent $\kappa$, that approximate
minimax estimation. Its central principle is to prioritize state--action pairs that are intrinsically difficult to estimate and insufficiently visited, using a $\kappa$-powered complexity-to-visitation ratio as the exploration reward.
However, implementing this principle under unknown dynamics requires estimating both the transition kernel and the intrinsic complexity, together with confidence bounds for these quantities. 
The resulting optimistic planning step jointly optimizes over a confidence set of transition kernels and their associated occupancy measures, introducing a computational burden that limits its practical applicability.

In this work, we develop Bayesian $\kappa$-Explorer by placing
independent Dirichlet priors on the transition distributions.
The posterior mean minimizes the posterior expected squared
$\ell_2$ estimation error, and the resulting minimum equals
the total posterior variance. Both admit closed-form expressions
in terms of the Dirichlet posterior parameters.
The posterior variance retains the intrinsic--extrinsic structure:
its numerator captures a posterior-mean analogue of transition
complexity, while its denominator incorporates the prior
concentration and the sampling effort accumulated under past
policies. This characterization provides a directly computable
exploration signal without separate confidence bounds on
intrinsic complexity.
This approach avoids constructing confidence sets for planning and replaces joint optimization over transition kernels and occupancy measures with planning under a single model, while retaining the same convergence rate.
To the best of our knowledge, this is the first
posterior-variance-driven exploration algorithm that combines
posterior-mean planning with a quantitative guarantee on the
maximum posterior expected squared transition-estimation error relative to the optimal asymptotic minimax allocation benchmark. 

The remainder of the paper is organized as follows. After formulating the problem, we present Bayesian $\kappa$-Explorer and state its guarantee on the maximum posterior variance. We then evaluate the algorithm on three benchmark environments, where it outperforms the tested baselines in worst-case
transition-estimation error. Finally, we develop the theoretical analysis and establish the guarantee by bounding its error components.


\section{Problem Formulation}
\label{sec:problem_formulation}

We consider a finite controlled Markov chain with state space
$\mathcal S$, $|\mathcal S|=S$, and action space
$\mathcal A$, $|\mathcal A|=A$.
At each time $t=0,1,\ldots$, the agent observes
$s_t\in\mathcal S$, selects $a_t\in\mathcal A$, and transitions
to $s_{t+1}$ according to an unknown kernel
$P^\star(\cdot\mid s_t,a_t)$.
A stationary randomized policy
$\pi:\mathcal S\to\Delta(\mathcal A)$ specifies the action
distribution at each state, and $\Pi$ denotes the set of
such policies.

Our goal is to estimate the entire transition kernel using $T$ observations.
We minimize the worst-case expected squared $\ell_2$ transition-estimation error:
\begin{equation}
\label{eq:minimax_error}
\max_{(s,a)\in\mathcal S\times\mathcal A}
\mathbb E_\pi\!\left[
\left\|
P^\star(\cdot\mid s,a)
-
\widehat P_T(\cdot\mid s,a)
\right\|_2^2
\right],
\end{equation}
where the expectation is over the trajectory generated by
$\pi$, and
$\widehat P_T$ is an estimator of the transition kernel.
To estimate the transition probabilities, we define the
transition count after $t$ samples as
\(
N_t(s,a,s')
:=
\sum_{\tau=0}^{t-1}
\mathbf 1\{s_\tau=s,\ a_\tau=a,\ s_{\tau+1}=s'\}
\)
and the state--action visit count as
\(
N_t(s,a):=\sum_{s'\in\mathcal S}N_t(s,a,s').
\)
Thus, $\sum_{s,a}N_t(s,a)=t$.
Let $T$ be the total sampling budget.

We adopt a Bayesian perspective and regard
$P^\star$ as a realization of a random transition kernel $P$
with independent Dirichlet priors on its rows:
\(
  P(\cdot\mid s,a)
  \sim
  \operatorname{Dir}\!\bigl(\alpha_0(s,a,\cdot)\bigr),
  \quad
  \alpha_0(s,a,s')>0.
\)
A classical choice is the Jeffreys prior, $\operatorname{Dir}(1/2,\ldots,1/2)$, whose posterior mean yields the Krichevsky--Trofimov estimator, known for its universal coding guarantees~\citep{krichevsky1981performance}. 
We write
$\beta_0(s,a):=\sum_{s'}\alpha_0(s,a,s')$
for the prior concentration and
$\mu_0(s'\mid s,a):=\alpha_0(s,a,s')/\beta_0(s,a)$
for the prior mean.
Together, $\beta_0$ and $\mu_0$ specify the Dirichlet prior.
Let $\mathcal F_t:=\sigma(s_0,a_0,s_1,\ldots,a_{t-1},s_t)$
denote the observed history after $t$ transitions.

By Dirichlet conjugacy,
\begin{equation}
    P(\cdot\mid s,a)\mid\mathcal F_t
      \sim
      \operatorname{Dir}\!\bigl(\alpha_t(s,a,\cdot)\bigr),
      \qquad
    \alpha_t(s,a,s')
      :=\alpha_0(s,a,s')+N_t(s,a,s').
  \label{eq:dirichlet-posterior}
\end{equation}
The posterior concentration and posterior mean are
\begin{equation}
    \beta_t(s,a)
      :=\sum_{s'}\alpha_t(s,a,s')
        =\beta_0(s,a)+N_t(s,a),\qquad
    \mu_t(s'\mid s,a)
      :=
      \frac{\alpha_t(s,a,s')}
           {\beta_t(s,a)}.
  \label{eq:beta-mu}
\end{equation}
Since we are interested in minimizing the MSE in \eqref{eq:minimax_error}, we use the posterior mean as our model estimator, setting $\widehat P_t:=\mu_t$. It is the minimum mean squared error (MMSE) estimator under the Bayesian model.
The posterior mean squared error of this estimator can be expressed using the standard Dirichlet variance formula~\citep{kotz2019continuous}:
\begin{equation}
  \begin{aligned}
    V_t(s,a)
      &:=
      \mathbb E\!\left[
        \left\|P(\cdot\mid s,a)-\mu_t(\cdot\mid s,a)\right\|_2^2
        \,\middle|\,\mathcal F_t
      \right]\\
      &=\sum_{s' \in \mathcal{S}}\operatorname{Var}\!\left[
        P(s'\mid s,a)\,\middle|\,\mathcal F_t
        \right]\\
        &=
        \sum_{s' \in \mathcal{S}}\frac{\mu_t(s'\mid s,a)
        \bigl(1-\mu_t(s'\mid s,a)\bigr)}
        {\beta_t(s,a)+1}\\
        &=\frac{c_t(s,a)}{\beta_t(s,a)+1},
  \end{aligned}
  \label{eq:total-posterior-variance}
\end{equation}
where $c_t(s,a):=1-\sum_{s'\in\mathcal S}\mu_t(s'\mid s,a)^2.$
This expression reveals an intrinsic--extrinsic decomposition
of posterior uncertainty.
The numerator $c_t(s,a)$ is a posterior-mean analogue of
the \emph{intrinsic complexity}
\begin{equation}
\label{eq:c_def}
c^\star(s,a)
:=
\sum_{s'\in\mathcal S}
P^\star(s'\mid s,a)
\bigl(1-P^\star(s'\mid s,a)\bigr),
\end{equation}
which characterizes the statistical difficulty of estimating
the next-state distribution at $(s,a)$.
Throughout the theoretical analysis, we assume
$c^\star(s,a)>0$ for every $(s,a)$.
The denominator $\beta_t(s,a)+1=\beta_0(s,a)+N_t(s,a)+1$ combines the prior concentration with the \emph{extrinsic} sampling effort. These two intrinsic and extrinsic factors together govern estimation accuracy.

\begin{remark}
In the frequentist setting considered in~\cite{gu2026kappa},
the empirical estimator based on a fixed number $n\ge1$ of
independent samples from $P^\star(\cdot\mid s,a)$ has
expected squared $\ell_2$ error $c^\star(s,a)/n$.
Equation~\eqref{eq:total-posterior-variance} retains this
intrinsic--extrinsic structure, with a prior-adjusted sample
count in the denominator. Unlike $c^\star(s,a)$, which depends
on the unknown kernel $P^\star$, the posterior coefficient
$c_t(s,a)$ is directly computable from $\mu_t$.
This provides an exploration signal without any need for a separate
confidence bound on intrinsic complexity. Combined with
planning under the posterior mean model, our approach avoids
constructing confidence sets for planning and jointly
optimizing over plausible transition kernels and their
associated occupancy measures.
\end{remark}

We evaluate the learned model by its worst-case posterior
uncertainty after $T$ observations:
\begin{equation}
    M_T
    :=
    \max_{(s,a)\in\mathcal S\times\mathcal A}V_T(s,a).
    \label{eq:minimax_var}
\end{equation}
Our goal is to design an adaptive sampling strategy that
makes $M_T$ as small as possible.
In the next section, we introduce an algorithm that uses
the posterior mean $\mu_t$ and posterior uncertainty $V_t$
to guide exploration toward reducing the maximum posterior
uncertainty.
It avoids constructing separate confidence bounds for the
transition kernel and its intrinsic complexity, thereby
simplifying the planning problem.

\section{Bayesian $\kappa$-Explorer}
\label{sec:method}

We introduce Bayesian $\kappa$-Explorer and establish an upper bound on the maximum posterior variance $M_T$ attained by the algorithm.

\subsection{Algorithm}
\label{subsec:algorithm}

Bayesian $\kappa$-Explorer takes as input the priors
$\alpha_0$, curvature parameter $\kappa\ge1$,
occupancy floor $\eta>0$, confidence level $\delta$, and episode lengths
$\tau_1,\ldots,\tau_J$.
At the start of each episode, the algorithm updates the
posterior using the observations collected so far, constructs
an exploration reward from the posterior uncertainty, and
plans under the posterior mean transition model.
The resulting policy is executed for the duration of the episode.

\paragraph{Exploration Reward.}
Let $t_m$ denote the start time of episode $m$.
At time $t_m$, the algorithm uses the accumulated counts to
compute the posterior mean $\mu_{t_m}$ and posterior variance
$V_{t_m}$, and assigns each state--action pair the exploration reward
\begin{equation}
  r_m(s,a):=V_{t_m}(s,a)^\kappa,\qquad \kappa\ge1.
  \label{eq:bayesian-kappa-reward}
\end{equation}
Increasing $\kappa$ amplifies the reward ratio between pairs
with higher and lower positive posterior variance, placing
greater emphasis on the most uncertain state--action pairs.

\paragraph{Posterior-Mean Planning.}
Given the exploration reward $r_m$, we plan under the
posterior mean transition kernel $\mu_{t_m}$.
To account for the dynamical constraints, define the
stationary occupancy domain for a transition kernel $p$ as
\begin{equation}
\label{eq:occupancy-set}
\mathcal D^{(p)}
:=
\left\{
\mathbf d\in\Delta(\mathcal S\times\mathcal A):
  \sum_{a\in\mathcal A}d(s,a)
  =
  \sum_{s',a'}p(s\mid s',a')d(s',a'),
  \qquad\forall s\in\mathcal S
\right\}.
\end{equation}
To assign strictly positive mass to every state--action pair in the
planned allocation, we introduce the constrained occupancy
domain for $\eta\in(0,1/(2SA))$:
\begin{equation}
\label{eq:restricted-occupancy-set}
\mathcal D_\eta^{(p)}
:=
\left\{
\mathbf d\in\mathcal D^{(p)}:
d(s,a)\ge2\eta,\quad
\forall(s,a)\in\mathcal S\times\mathcal A
\right\}.
\end{equation}
We assume that $\eta$ is chosen small enough such that $\mathcal D_\eta^{(\mu_{t_m})}$ is nonempty
at every episode.
The planner computes the episode policy through
\begin{equation}
  \boldsymbol{\psi}_m
  \in
  \operatorname*{arg\,max}_{
    \mathbf d\in\mathcal D_\eta^{(\mu_{t_m})}}
  \sum_{s,a}r_m(s,a)d(s,a),
  \qquad
  \pi_m(a\mid s)
  :=
  \frac{\psi_m(s,a)}{\sum_{b\in\mathcal A}\psi_m(s,b)}.
  \label{eq:bayesian-kappa-oracle}
\end{equation}
The algorithm executes $\pi_m$ on the true MDP for $\tau_m$
transitions, starting from the current state, and updates
the transition counts using the resulting observations.
After the final episode, it returns the posterior mean
$\mu_T$, where $T=\sum_{m=1}^J\tau_m$.
Algorithm~\ref{alg:bayesian-kappa-explorer} summarizes
these steps.

\begin{algorithm2e}[H]
  \caption{Bayesian $\kappa$-Explorer}
  \label{alg:bayesian-kappa-explorer}
  \KwIn{Priors $\alpha_0$, curvature parameter $\kappa\ge1$,
        occupancy lower-bound parameter $\eta>0$, confidence level $\delta$, episode lengths $\tau_1,\ldots,\tau_J$}
  \KwOut{Transition estimate $\widehat P_T$}
  Set $t_1\gets0$ and initialize all transition counts to zero\;
  \For{$m=1,\ldots,J$}{
    Compute $\mu_{t_m}$ and $V_{t_m}$ using
    \eqref{eq:beta-mu} and \eqref{eq:total-posterior-variance}\;
    Set $r_m(s,a)\gets V_{t_m}(s,a)^\kappa$ for each $(s,a)$\;
    Solve \eqref{eq:bayesian-kappa-oracle} for $\boldsymbol{\psi}_m$
    and form $\pi_m$\;
    Execute $\pi_m$ for $\tau_m$ transitions and update the counts\;
    Set $t_{m+1}\gets t_m+\tau_m$\;
  }
  Compute $\mu_T$ from the final counts\;
  \textbf{return} $\widehat P_T=\mu_T$\;
\end{algorithm2e}

\subsection{Main Guarantee}
\label{subsec:main-guarantee}

To benchmark $M_T$, we consider the best allocation in the
constrained occupancy domain of the true kernel:
\begin{equation}
  C^\star
  :=
  \min_{\mathbf d\in\mathcal D^{(P^\star)}}\max_{(s,a)\in\mathcal S\times\mathcal A}   \frac{c^\star(s,a)}{d(s,a)},
  \qquad
  \mathbf d^\star
  \in
  \operatorname*{arg\,min}_{
    \mathbf d\in\mathcal D^{(P^\star)}}
  \max_{(s,a)\in\mathcal S\times\mathcal A}\frac{c^\star(s,a)}{d(s,a)},
  \label{eq:minimax-allocation-benchmark}
\end{equation}
where $\max_{(s,a)\in\mathcal S\times\mathcal A} \frac{c^\star(s,a)}{d(s,a)}:=+\infty$ if any component of
$\mathbf d$ is zero.
The corresponding asymptotic variance benchmark is
$C^\star/T$.
We can now state the main guarantee.

\begin{theorem}[Maximum posterior variance bound]
\label{thm:bayesian-kappa-main}
Under technical Assumptions~\ref{ass:communicating-mdp}
and~\ref{assumption:stability}, stated below, run
Algorithm~\ref{alg:bayesian-kappa-explorer} with $\kappa\ge1$
and appropriately chosen $\eta$ and episode lengths $\tau_m$.
For every $\delta\in(0,1)$, there exists a threshold
$T_0(\delta)$ such that, for every $T\ge T_0(\delta)$,
the following bounds hold with probability at least $1-\delta$.
\begin{equation}
    M_T\le{}\frac{C^\star}{T}
    +\underbrace{\frac{(\eta^{-1/\kappa}-1)C^\star}{T}}
        _{\text{finite-$\kappa$ approximation}}
    +\underbrace{\frac{\eta^{-1/\kappa}(\kappa-1)
                       \varepsilon_{\mathrm{opt}}(T,\delta/2)}
                       {\kappa T(C^\star)^{\kappa-1}}}
        _{\text{allocation optimization}}
    +\underbrace{\varepsilon_{\mathrm{post}}(T,\delta/2)}
        _{\text{posterior estimation}}.
  \label{eq:bayesian-kappa-expanded-bound}
\end{equation}
On the same event, the following bound applies for all
$C^\star\ge0$:
\begin{equation}
  M_T\le
  \frac{\eta^{-1/\kappa}}{T}
  \left[(C^\star)^\kappa+
        (\kappa-1)\varepsilon_{\mathrm{opt}}(T,\delta/2)\right]^{1/\kappa}
  +\varepsilon_{\mathrm{post}}(T,\delta/2).
  \label{eq:bayesian-kappa-main-bound}
\end{equation}
\end{theorem}

The finite-$\kappa$ term in \eqref{eq:bayesian-kappa-expanded-bound} comes from approximating the maximum in \eqref{eq:minimax-allocation-benchmark} by aggregating the allocation costs across state--action pairs. For fixed $\eta$, the approximation factor $\eta^{-1/\kappa}$ tends to one as
$\kappa$ grows.
The allocation optimization term reflects the surrogate objective gap between the empirical allocation and the optimal allocation. Our analysis yields an upper bound
$\varepsilon_{\mathrm{opt}}(T,\delta)
=\widetilde{\mathcal O}(T^{-1/3})$,
so its contribution to the posterior variance bound is
$\widetilde{\mathcal O}(T^{-4/3})$
for fixed $\kappa$, $\eta$, $\delta$, and $C^\star$.
The posterior estimation term
$\varepsilon_{\mathrm{post}}(T,\delta)$
controls the contribution of the coefficient error. Under the visitation floor, this term is $\widetilde{\mathcal O}(T^{-3/2})$ for fixed $\eta$ and $\delta$.
Section~\ref{sec:analysis} provides the technical assumptions, the specific choice of $\eta$ and $\tau_m$, as well as the details of the analysis.

\section{Experiments}
\label{sec:experiments}

We evaluate how effectively the proposed algorithm reduces the estimation error on three discretized domains compared with other methods.

\subsection{Experimental Setup}
\label{subsec:experimental-setup}

\paragraph{Environments.}
We evaluate transition model estimation in three tabular MDPs based on
Pendulum, Mountain Car, and Taxi from the Gymnasium control suite introduced by~\cite{brockman2016openai}. In \textbf{Pendulum}, we discretize the
angle and angular velocity into ten bins each and use five torque levels,
giving $S=100$ states and $A=5$ actions. In \textbf{Mountain Car}, we discretize
position and velocity into sixteen and seven bins, respectively, and
retain the three original actions, giving $S=112$ and $A=3$.
We introduce stochastic transitions by perturbing the updated angle
and angular velocity in Pendulum and the updated velocity in Mountain Car.
In \textbf{Taxi}, we use a variant of the original $5\times5$ map with
$S=400$ states and $A=6$ actions. Its action perturbations act through
legal Taxi transitions and respect the walls of the map.

\paragraph{Baselines.}
We compare Bayesian $\kappa$-Explorer with the following exploration methods.
\textbf{Random} selects an action uniformly at random at each time step, $A_t \sim \mathcal{U}(\mathcal{A})$.
\textbf{MaxEnt}~\citep{hazan2019provably} maximizes visitation entropy to encourage broad exploration. We use the joint state--action visitation distribution so that the objective encourages coverage of both states and actions.
\textbf{ModEst\_LSE}~\citep{tarbouriech2020active} optimizes a log-sum-exp surrogate for the worst-case transition estimation error.
\textbf{Weighted\_MaxEnt}~\citep{tarbouriech2020active} maximizes weighted state--action visitation entropy, assigning greater weight to transitions that are harder to estimate.
\textbf{State Marginal Matching (SMM)}~\citep{lee2019efficient} encourages the visitation distribution to match a target. We adapt SMM to state--action pairs and set the target mass proportional to the estimated transition complexity of each pair.

\paragraph{Evaluation protocol.}
We test each method with 20 random seeds in each environment. 
The sampling budget is the number of observed
transitions $T$, with a maximum of $100{,}000$ for Pendulum and Taxi and
$300{,}000$ for Mountain Car. We evaluate the transition estimate
$\widehat P_t$ at each checkpoint using the worst-case $\ell_2$ error
\[
    \max_{(s,a)\in\mathcal S\times\mathcal A}
    \left\|
        \widehat P_t(\cdot\mid s,a)
        - P^\star(\cdot\mid s,a)
    \right\|_2 .
\]
We take the maximum over state--action pairs separately for each seed
and then average the resulting errors across seeds.
For Taxi, $P^\star$ is the exact transition kernel.
For Pendulum and Mountain Car, we construct a fixed reference kernel
from independent simulator samples for each state--action pair.
The reference kernels are used only for evaluation.

\subsection{Pendulum Case Study}
\label{subsec:pendulum-case-study}

\begin{figure}[t]
        {%
            \hfill
            \subfigure[Pendulum]{
            \raisebox{1.2cm}{%
                \includegraphics[width=0.2\linewidth]{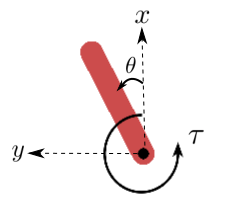}%
                \label{fig:pendulum_illustration}
            }}
            \hfill
            \subfigure[Worst-case error vs. budget ($S = 100, A = 5$)]{%
                \includegraphics[width=0.6\linewidth]{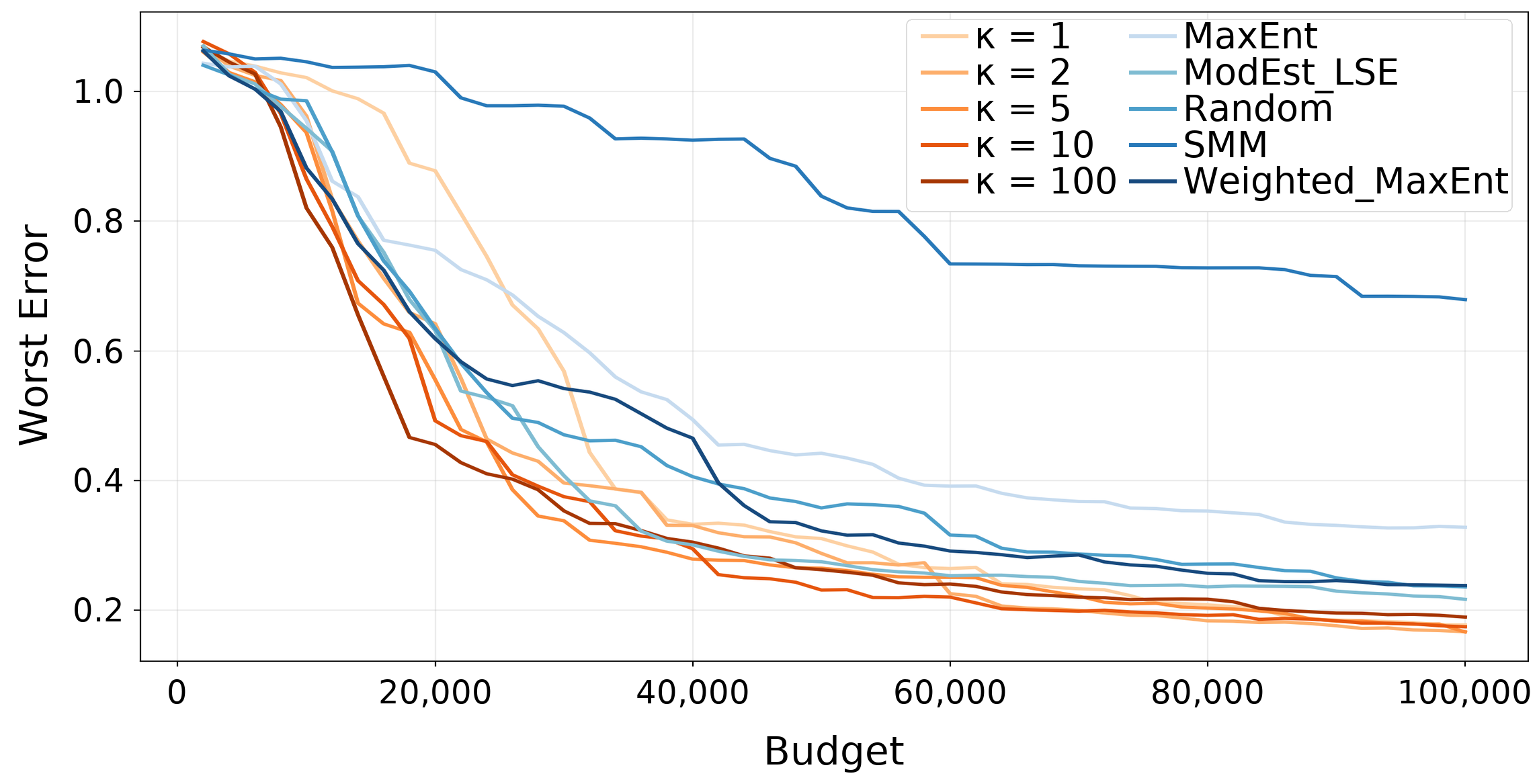}%
                \label{fig:pendulum_error}
            }%
        }
        \caption{\textbf{Pendulum: }Worst-case transition estimation error versus sampling budget. Orange curves correspond to Bayesian $\kappa$-Explorer with $\kappa \in \{1,2,5,10,100\}$, and blue curves correspond to the baselines. Each curve is averaged over 20 random seeds.}
        \label{fig:pendulum}
\end{figure}

Figure~\ref{fig:pendulum_illustration} depicts the Pendulum environment whose dynamics remains challenging to estimate even after every state--action pair has been visited. Under random actions, the pendulum frequently passes through the lower part of its swing, whereas near-upright states with low angular velocity are harder to revisit or remain at. Consequently, some of the states remain poorly sampled and dominate the worst-case error.

$\kappa$-Explorer addresses this imbalance through its exploration
reward. Among pairs with comparable intrinsic complexities,
the score prioritizes those with fewer samples. The planner can then
choose a sequence of torques that brings the pendulum back to the
desired angle and angular velocity and samples the action of interest.
In Figure~\ref{fig:pendulum_error}, several choices of $\kappa$ reduce
the error faster than the strongest baselines during the early part
of data collection. At the largest sampling budget, all settings of $\kappa$-Explorer achieve a lower worst-case error than the
best-performing baseline.

The effect of $\kappa$ is nonmonotonic. Moderate values can reduce the
error faster or achieve a lower final error than $\kappa=100$.
Increasing $\kappa$ concentrates the exploration reward on the
highest-scoring pairs. For example, if those pairs lie near the upright
position, a very large $\kappa$ may favor trajectories that use large
torques ($+2/-2$) to reach them quickly. On the other hand, transitions under smaller torques ($+1/-1$) along the way may then be neglected and remain undersampled. This illustrates a possible mechanism for the observed trade-off. A moderate $\kappa$ gives more
weight to these additional sampling opportunities, which can help
reduce the worst-case error across all state--action pairs.

\subsection{Results on Mountain Car and Taxi}
\label{subsec:mountaincar-taxi-results}

\begin{figure}
    \centering
    \includegraphics[width=0.98\linewidth]{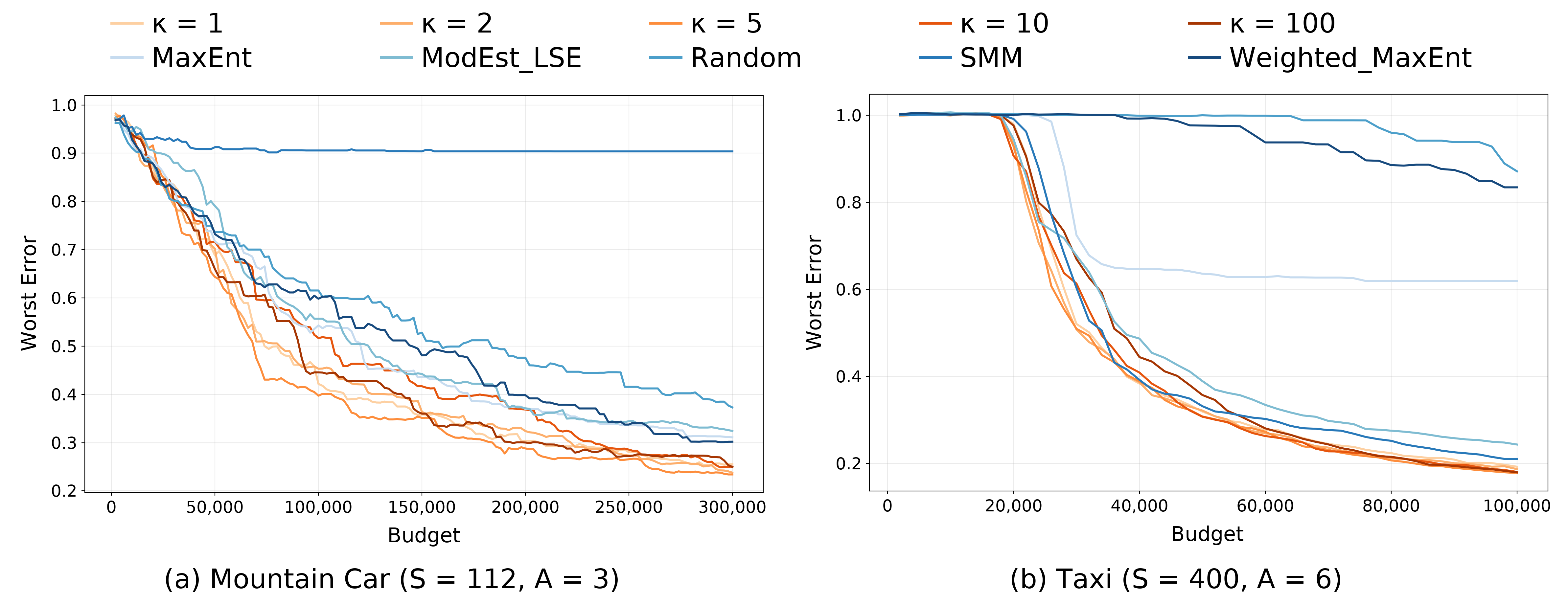}
    \caption{\textbf{Mountain Car and Taxi: }Worst-case transition estimation error versus sampling budget. Orange curves show Bayesian $\kappa$-Explorer for $\kappa \in \{1,2,5,10,100\}$, and blue curves show the baselines. Each curve is averaged over 20 random seeds.}
        \label{fig:error_comp}
\end{figure}

Figure~\ref{fig:error_comp} compares Bayesian $\kappa$-Explorer with the baselines on Mountain Car and Taxi. In both environments, all tested values of
$\kappa$ achieve lower worst-case transition estimation errors than
the baselines at larger sampling budgets. On Mountain Car, moderate
values of $\kappa$ reduce the error faster than $\kappa=100$ over parts
of the budget range. On Taxi, the $\kappa$-Explorer curves are closer together toward the end of the run. Overall, $\kappa$-Explorer outperforms the baselines at larger sampling budgets in both environments, while the sensitivity to $\kappa$ differs between them.

\section{Theoretical Analysis}
\label{sec:analysis}

We prove Theorem~\ref{thm:bayesian-kappa-main} in three steps.
First, we relate the minimax allocation objective to the smooth surrogate $U_\kappa$. Next, we bound the convergence gap of Bayesian $\kappa$-Explorer on $U_\kappa$. Finally, we bound the posterior estimation error and combine the results.

\subsection{Surrogate Objective and Minimax Approximation}
\label{subsec:utility}
\label{subsec:finite-kappa}

For strictly positive allocations $\mathbf{d} \in \Delta(\mathcal{S} \times \mathcal{A})$ and curvature parameter $\kappa \in \mathbb Z^+$, define the concave utility
\begin{equation}
  U_\kappa(\mathbf d):=
  \begin{cases}
    \displaystyle\sum_{s,a}c^\star(s,a)\log d(s,a), & \kappa=1,\\[3pt]
    \dfrac{1}{1-\kappa}\displaystyle\sum_{s,a}
      c^\star(s,a)^\kappa d(s,a)^{1-\kappa}, & \kappa>1.
  \end{cases}
  \label{eq:kappa-utility}
\end{equation}
The objective $U_\kappa$ is concave and separable across state--action
pairs. A key property is that its gradient satisfies
\(
\bigl(\nabla U_\kappa(\mathbf d)\bigr)_{s,a}
=
\left(
\frac{c^\star_{s,a}}{d(s,a)}
\right)^\kappa.
\)

\begin{proposition}[Finite-$\kappa$ approximation]
\label{prop:kappa-approximation}
Let $\kappa>1$. For any probability vector $\mathbf d$ with
$d(s,a)\ge\eta$ for all $(s,a)$,
\begin{equation}
  \eta^{1/\kappa} \max_{s, a}\frac{c^\star(s, a)}{d(s, a)}
  \le \bigl[(1-\kappa)U_\kappa(d)\bigr]^{1/\kappa}
  \le \max_{s, a}\frac{c^\star(s, a)}{d(s, a)}.
  \label{eq:kappa-approximation}
\end{equation}
\end{proposition}
The proof is the same as that of Proposition~1 in \cite{gu2026kappa}.

\subsection{Convergence Analysis}
\label{subsec:optimization}

We analyze the convergence of
Algorithm~\ref{alg:bayesian-kappa-explorer} through a
Frank--Wolfe \citep{frank1956algorithm} interpretation.
For $t\ge1$, define the empirical visitation vector
$\widehat{\mathbf d}_t$ by its components
\(
\widehat d_t(s,a):=N_t(s,a)/t.
\)
For strictly positive $\widehat{\mathbf d}_{t_m}$, the
rescaled reward approximates
$\nabla U_\kappa(\widehat{\mathbf d}_{t_m})$ by replacing
$c^\star(s,a)$ with $c_{t_m}(s,a)$ and incorporating the
prior-dependent offset in the denominator.
Since multiplication by the positive scalar $t_m^\kappa$
leaves the maximizers of
\eqref{eq:bayesian-kappa-oracle} unchanged, the planning
step can be interpreted as an approximate Frank--Wolfe
linear maximization step, using an estimated gradient
and a feasible occupancy domain determined by the
posterior mean kernel.
For the analysis, we impose the following communicating
assumption.
\begin{assumption}[Communicating MDP]
\label{ass:communicating-mdp}
The finite MDP $\mathcal M^\star=(\mathcal S,\mathcal A,P^\star)$
is communicating: for every pair of states $s,s'$, there
exist a policy $\pi$ and a finite time $t\ge0$ such that
\[
  \mathbb P_{P^\star}^\pi(S_t=s'\mid S_0=s)>0.
\]
\end{assumption}
With that, we establish the following convergence guarantees.
\begin{lemma}[Surrogate convergence rate]
\label{lem:general_convergence_rate}
Let $P^\star$ satisfy Assumption~\ref{ass:communicating-mdp}. Fix $\kappa\ge1$ and let the episode lengths satisfy
\(
\tau_m=\tau_1m^2,
\)
such that 
\(
t_m=\sum_{j=1}^{m-1}\tau_j
  =\tau_1\frac{(m-1)m(2m-1)}{6}.
\)
Let $\mathbf d_\kappa^\star \in \arg\max_{\mathbf d \in \mathcal{D}^{(P^\star)}} U_\kappa(\mathbf d)$ be the optimizer.
Assume that $\eta$ is chosen such that
\(
2\eta
<
\min_{s,a}d_{\kappa}^\star(s, a).
\)
Such an $\eta$ exists by Lemma~\ref{lem:positive-minimax-occupancy}
in Appendix~\ref{Appendix:Positivefeasibleoccupancy}.
Then, after $J$ episodes, with $T=\sum_{m=1}^J\tau_m$, we have, with probability at least $1-\delta$,
\[ 
\varepsilon_{\mathrm{opt}}(T,\delta)
:=U_\kappa(\mathbf d_\kappa^\star)
-
U_\kappa\!\left(\widehat{\mathbf d}_T\right)
\le 
\widetilde{\mathcal O}\!\left(T^{-1/3}\right).
\]
\end{lemma}
The proof of Lemma~\ref{lem:general_convergence_rate} follows the same structure as that of Theorem~2
in \cite{gu2026kappa}, except for the method used to bound
\(
\epsilon_m
:=
\left\langle
\nabla U_\kappa(\widehat{\mathbf d}_{t_m}),
\boldsymbol{\psi}_m^\star-\boldsymbol{\psi}_m
\right\rangle,
\)
which captures the optimization and estimation error induced
by planning under unknown transition dynamics.
For $t_m>0$, define the normalized planning weights by
\(
  w_m(s,a):=[t_mV_{t_m}(s,a)]^\kappa.
\)
This positive rescaling leaves the planning maximizers
unchanged.
Define the occupancies selected by the posterior-mean planner
and the true-model gradient oracle, together with their
weighted discrepancy, as follows:
\[
  \boldsymbol{\psi}_m
  \in
  \operatorname*{arg\,max}_{
    \mathbf d\in\mathcal D_\eta^{(\mu_{t_m})}}
  \langle\mathbf w_m,\mathbf d\rangle,\qquad
  \boldsymbol{\psi}_m^\star
  \in
  \operatorname*{arg\,max}_{
    \mathbf d\in\mathcal D_\eta^{(P^\star)}}
  \left\langle
    \nabla U_\kappa(\widehat{\mathbf d}_{t_m}),\mathbf d
  \right\rangle,
  \qquad
  G_m
  :=
  \left\langle
    \mathbf w_m,
    \boldsymbol{\psi}_m^\star-\boldsymbol{\psi}_m
  \right\rangle.
\]
In the frequentist analysis (Appendix~B of \cite{gu2026kappa}),
the analogous term $G_m$ is nonpositive on the event
$P^\star\in\mathcal B_{t_m}$,
because the true-model oracle occupancy $\boldsymbol{\psi}_m^\star$ is feasible for the optimistic planner and both are evaluated using the same planning weights.
In contrast, in the Bayesian setting, $\boldsymbol{\psi}_m^\star$ need not belong to the posterior-mean planning domain
$\mathcal D_\eta^{(\mu_{t_m})}$.

We define the transition estimation error by
\(
  \Delta_m^P
  :=
  \max_{s,a}
  \left\|
    P^\star(\cdot\mid s,a)
    -
    \mu_{t_m}(\cdot\mid s,a)
  \right\|_1.
\)
We bound the resulting discrepancy using the following stability assumption with respect to the transition estimation error.
\begin{assumption}[Uniform occupancy stability]
\label{assumption:stability}
There exists a constant $L<\infty$, independent of the
policy and episode, such that for every stationary policy
$\pi$ with $\pi(a\mid s)>0$ for all $(s,a)$ and every
episode $m$, the induced chains under $P^\star$ and
$\mu_{t_m}$ admit unique stationary occupancy measures
satisfying
\(
  \left\|
    \mathbf d_{\mu_{t_m}}^\pi
    -
    \mathbf d_{P^\star}^\pi
  \right\|_1
  \le L\Delta_m^P.
\)
\end{assumption}
Assumption~\ref{assumption:stability} ensures that, for a
fixed policy, the stationary occupancy measure is stable
in $\ell_1$ distance under small perturbations of the
transition kernel.
Together with strict feasibility under the true kernel,
this implies that the true-model oracle occupancy
$\boldsymbol{\psi}_m^\star$ admits a feasible counterpart
$\bar{\boldsymbol{\psi}}_m
\in\mathcal D_\eta^{(\mu_{t_m})}$ satisfying
\(
  \|\boldsymbol{\psi}_m^\star
    -\bar{\boldsymbol{\psi}}_m\|_1
  \le C_{\rm occ}\Delta_m^P,
\)
where $C_{\rm occ}$ is independent of the episode index $m$.
This result is established in Lemma~\ref{lem:feasible-occupancy-counterpart} in
Appendix~\ref{appendix:lem_feasible-occupancy-counterpart}.
Therefore, we can find a
$\bar{\boldsymbol{\psi}}_m\in\mathcal D_\eta^{(\mu_{t_m})}$
that is close to $\boldsymbol{\psi}_m^\star$ and also satisfies
$\langle\mathbf w_m,\boldsymbol{\psi}_m\rangle
\ge \langle\mathbf w_m,\bar{\boldsymbol{\psi}}_m\rangle$,
which allows us to write
\[
G_m
  =
  \langle\mathbf w_m,
    \boldsymbol{\psi}_m^\star
    -\bar{\boldsymbol{\psi}}_m\rangle
  +
  \underbrace{
    \langle\mathbf w_m,
      \bar{\boldsymbol{\psi}}_m
      -\boldsymbol{\psi}_m\rangle
  }_{\le0}.
\]
Further bounding the first term gives an upper bound of
$O(t_m^{-1/2})$ on $G_m$; details are provided in
Appendix~\ref{appendix:Prop_Posterior_bound_planning_discrepancy}.
We further conclude that, at each fixed episode and for fixed posterior failure probability $\delta$, $\epsilon_m$ admits an $O(t_m^{-1/2})$ upper bound, with all other parameters
held constant (see Appendix~\ref{appendix:rest_epsilon_bound} for details).
This matches the $O(t_m^{-1/2})$ rate of the corresponding frequentist oracle-error bound, while avoiding the construction of a confidence ball for planning.

\subsection{Posterior Coefficient Error}
\label{subsec:posterior-comparison}

We next bound the remaining term, which accounts for the discrepancy
between the posterior and true intrinsic complexities.

\begin{lemma}[Posterior coefficient error]
\label{lem:posterior-coefficient-error}
For $T\ge1$ and $\delta\in(0,1)$, let
$q_T(\delta):=\log(4S^2AT/\delta)$ and
$\beta_{\max}:=\max_{s,a}\beta_0(s,a)$. With probability at least
$1-\delta/2$ under the fixed-kernel trajectory law, if
$\widehat d_T(s,a)\ge\eta$ for all $(s,a)$, then
\begin{equation}
\begin{aligned}
    \varepsilon_{\mathrm{post}}(T,\delta)
  &:=
  \max_{s,a}\frac{|c_T(s,a)-c^\star(s,a)|}
                    {N_T(s,a)+\beta_0(s,a)+1}\\
  &\le \frac{1}{\eta T+\min_{s,a}\beta_0(s,a)+1}
  \left(
    \sqrt{\frac{2S q_T(\delta)}{\eta T}}
    +\frac{2\sqrt{2}\,\beta_{\max}}{\eta T}
  \right).
  \label{eq:bayesian-kappa-posterior-error}
  \end{aligned}
\end{equation}
\end{lemma}
The proof of Lemma~\ref{lem:posterior-coefficient-error}
is in Appendix~\ref{appendix:proof_lem:posterior-coefficient-error}.
Combining the finite-$\kappa$ approximation bound from Proposition~\ref{prop:kappa-approximation}, the optimization error bound $\varepsilon_{\mathrm{opt}}(T,\delta)$ from Lemma~\ref{lem:general_convergence_rate}, and the posterior coefficient error bound $\varepsilon_{\mathrm{post}}(T,\delta)$ from Lemma~\ref{lem:posterior-coefficient-error} yields Theorem~\ref{thm:bayesian-kappa-main}.
The proof is provided in Appendix~\ref{appendix:proof_thm_bayesian-kappa-main}.





\section{Conclusion}
\label{sec:conclusion}
In this work, we build on the frequentist $\kappa$-Explorer algorithm introduced in~\cite{gu2026kappa} and develop a Bayesian approach to minimax transition estimation in unknown MDPs. We retain the parameterized family of exploration objectives that approximate minimax estimation and establish convergence through a Frank--Wolfe interpretation. By placing Dirichlet priors on the transition distributions, we express posterior expected squared estimation error as a closed-form posterior variance and use its $\kappa$th power to guide exploration. Planning under the posterior mean model eliminates separate confidence-bound constructions and joint optimization over transition kernels and occupancy measures, yielding a simpler and computationally more efficient algorithm. Under the stated assumptions, we preserve the $\widetilde{\mathcal O}(T^{-1/3})$ surrogate convergence rate of frequentist $\kappa$-Explorer and bound the maximum posterior variance relative to the optimal asymptotic minimax allocation benchmark. Experiments on Pendulum, Mountain Car, and Taxi demonstrate lower worst-case transition-estimation errors than the tested baselines, both early in data collection and at larger sampling budgets.

\newpage
\subsection*{Generative AI Use}
Generative AI tools were used to assist with revising and polishing the manuscript, developing portions of the proofs, and writing portions of the experimental code.
The authors reviewed and verified all AI-assisted content, including the mathematical arguments and experimental code, and conducted all experiments themselves.
The authors take full responsibility for the accuracy and integrity of the final manuscript.

\acks{We gratefully acknowledge support from ONR, the NSF TILOS
AI Institute, and the UCSD Centers for Machine Intelligence,
Computing, and Security (MICS).}

\bibliography{l4dc2027-sample}


\newpage

\appendix
\section{Positive Feasible Occupancy}
\label{Appendix:Positivefeasibleoccupancy}

\begin{lemma}[Positive feasible occupancy]
\label{lem:positive-minimax-occupancy}
Under Assumption~\ref{ass:communicating-mdp}, there exists
a stationary occupancy measure
$\bar{\mathbf d}\in\mathcal D^{(P^\star)}$ such that
\[
  d_{\min}:=\min_{s,a}\bar d(s,a)>0.
\]
Consequently, $\mathcal D_\eta^{(P^\star)}$ is nonempty
whenever $0<2\eta\le d_{\min}$.
\end{lemma}
The proof follows the same argument as that of
Lemma~1 in \cite{gu2026kappa}.

According to Lemma~\ref{lem:positive-minimax-occupancy}, a strictly positive feasible occupancy exists and $c^\star(s,a)>0$ for every $(s,a)$, thus any minimizer $\mathbf d^\star_\kappa$ must satisfy
\[
    \min_{s,a}d^\star_\kappa(s,a)>0.
\]
Consequently, there exists $\eta\in(0,1/(2SA))$
with $2\eta\le \min_{s,a}d^\star_\kappa(s,a)$ such that
$\mathbf d^\star_\kappa \in\mathcal D_\eta^{(P^\star)}$.

\section{Feasible Occupancy Counterpart}
\label{appendix:lem_feasible-occupancy-counterpart}
\begin{lemma}[Feasible occupancy counterpart]
\label{lem:feasible-occupancy-counterpart}
Suppose Assumption~\ref{assumption:stability} holds.
Assume that there exist a reference occupancy
$\bar{\mathbf d}\in\mathcal D^{(P^\star)}$ and a constant
$h>0$ such that
\[
  \bar d(s,a)\ge2\eta+h
  \qquad\text{for every }(s,a),
\]
and that $\mathcal D_\eta^{(\mu_{t_m})}$ is nonempty at
every episode.
Then, for every $\mathbf d\in\mathcal D_\eta^{(P^\star)}$
and every episode $m$, there exists
$\widetilde{\mathbf d}\in\mathcal D_\eta^{(\mu_{t_m})}$
such that
\[
  \|\widetilde{\mathbf d}-\mathbf d\|_1
  \le C_{\rm occ}\Delta_m^P,
  \qquad
  C_{\rm occ}:=L\left(1+\frac{2}{h}\right).
\]
In particular, the true-model oracle occupancy
$\boldsymbol{\psi}_m^\star$ admits a feasible counterpart
$\bar{\boldsymbol{\psi}}_m
\in\mathcal D_\eta^{(\mu_{t_m})}$ satisfying
\[
  \|\boldsymbol{\psi}_m^\star
    -\bar{\boldsymbol{\psi}}_m\|_1
  \le C_{\rm occ}\Delta_m^P.
\]

\end{lemma}
\begin{proof}
Fix an episode $m$ and an occupancy
$\mathbf d\in\mathcal D_\eta^{(P^\star)}$, and write
$e_m:=L\Delta_m^P$.
Let $\pi$ and $\bar\pi$ be the stationary policies induced
by $\mathbf d$ and $\bar{\mathbf d}$, respectively.
Both policies assign positive probability to every action
at every state.
Their stationary occupancy measures under the posterior
mean kernel are
\[
  \mathbf q:=\mathbf d_{\mu_{t_m}}^\pi,
  \qquad
  \bar{\mathbf q}:=\mathbf d_{\mu_{t_m}}^{\bar\pi}.
\]
By Assumption~\ref{assumption:stability},
\[
  \|\mathbf q-\mathbf d\|_1\le e_m,
  \qquad
  \|\bar{\mathbf q}-\bar{\mathbf d}\|_1\le e_m.
\]
Consequently, for every $(s,a)$,
\[
  q(s,a)\ge2\eta-e_m,
  \qquad
  \bar q(s,a)\ge2\eta+h-e_m.
\]

First suppose $e_m\le h$.
Although $\mathbf q$ need not satisfy the occupancy floor,
both $\mathbf q$ and $\bar{\mathbf q}$ belong to
$\mathcal D^{(\mu_{t_m})}$.
Since this domain is convex, the mixture
\[
  \widetilde{\mathbf d}
  :=
  (1-\lambda_m)\mathbf q+\lambda_m\bar{\mathbf q},
  \qquad
  \lambda_m:=\frac{e_m}{h}\in[0,1],
\]
also belongs to $\mathcal D^{(\mu_{t_m})}$.
Moreover,
\[
  \widetilde d(s,a)
  \ge
  (1-\lambda_m)(2\eta-e_m)
  +\lambda_m(2\eta+h-e_m)
  =2\eta.
\]
Thus, $\widetilde{\mathbf d}\in
\mathcal D_\eta^{(\mu_{t_m})}$.
Using the fact that the $\ell_1$ distance between two
probability vectors is at most $2$, we obtain
\begin{align*}
  \|\widetilde{\mathbf d}-\mathbf d\|_1
  &\le
  \|\mathbf q-\mathbf d\|_1
  +\lambda_m\|\bar{\mathbf q}-\mathbf q\|_1\\
  &\le e_m+\frac{2e_m}{h}\\
  &=C_{\rm occ}\Delta_m^P.
\end{align*}

Now suppose $e_m>h$.
By nonemptiness, choose any
$\widetilde{\mathbf d}\in\mathcal D_\eta^{(\mu_{t_m})}$.
Then
\[
  \|\widetilde{\mathbf d}-\mathbf d\|_1
  \le2
  <\frac{2e_m}{h}
  \le C_{\rm occ}\Delta_m^P.
\]
This proves the claim in both cases.
Taking $\mathbf d=\boldsymbol{\psi}_m^\star$ yields the
stated oracle counterpart.
\end{proof}

\section{Posterior Bound on the Planning Discrepancy}
\label{appendix:Prop_Posterior_bound_planning_discrepancy}

\begin{proposition}[Posterior bound on the planning discrepancy]
\label{proposition:Posterior_bound_planning_discrepancy}
Fix an episode $m$ with $t_m>0$.
Suppose $\mathbf w_m$ is $\mathcal F_{t_m}$-measurable,
$\|\mathbf w_m\|_\infty\le W$, and
$N_{t_m}(s,a)\ge\nu t_m$ for every $(s,a)$, where
$W,\nu>0$ are deterministic constants independent of $m$.

Define the event
\(
  \mathcal E_m
  :=
  \left\{
    \Delta_m^P
    \le
    \left(
      \frac{S}{\delta}
      \sum_{s,a}V_{t_m}(s,a)
    \right)^{1/2}
  \right\}.
\)
By the posterior variance identity,
\(
  \mathbb E\!\left[
    (\Delta_m^P)^2\mid\mathcal F_{t_m}
  \right]
  \le S\sum_{s,a}V_{t_m}(s,a).
\)
Let $\mathcal H_m$ be an event on which the conditions of
Lemma~\ref{lem:feasible-occupancy-counterpart} hold with
a common deterministic constant $C_{\rm occ}$. Conditional Markov's inequality therefore gives
\(
  \Pr(\mathcal E_m\mid\mathcal F_{t_m})
  \ge1-\delta.
\)

On $\mathcal E_m\cap\mathcal H_m$,
\begin{equation}
\begin{aligned}
  G_m^+
  &:=\max\{G_m,0\}\\
  &\le R_m
  :=C_{\rm occ}W
  \left(
    \frac{S}{\delta}
    \sum_{s,a}
    \frac{c_{t_m}(s,a)}{\beta_{t_m}(s,a)+1}
  \right)^{1/2}\\
  &\le
  C_{\rm occ}WS
  \sqrt{
    \frac{A}
    {\delta(\beta_{\min}+\nu t_m+1)}
  },
\end{aligned}
\label{eq:mean-gap-rate}
\end{equation}
where
$\beta_{\min}:=\min_{s,a}\beta_0(s,a)$.
For fixed $\delta$, the final bound is $O(t_m^{-1/2})$.
\end{proposition}

\begin{proof}
On $\mathcal H_m$,
Lemma~\ref{lem:feasible-occupancy-counterpart} provides
$\bar{\boldsymbol{\psi}}_m
\in\mathcal D_\eta^{(\mu_{t_m})}$ satisfying
\[
  \|\boldsymbol{\psi}_m^\star
    -\bar{\boldsymbol{\psi}}_m\|_1
  \le C_{\rm occ}\Delta_m^P.
\]
By the optimality of the posterior-mean planner,
\begin{align*}
  G_m
  &=
  \langle\mathbf w_m,
    \boldsymbol{\psi}_m^\star
    -\bar{\boldsymbol{\psi}}_m\rangle
  +
  \underbrace{
    \langle\mathbf w_m,
      \bar{\boldsymbol{\psi}}_m
      -\boldsymbol{\psi}_m\rangle
  }_{\le0}\\
  &\le
  W\|\boldsymbol{\psi}_m^\star
    -\bar{\boldsymbol{\psi}}_m\|_1\\
  &\le C_{\rm occ}W\Delta_m^P.
\end{align*}
Since the right-hand side is nonnegative, the same bound
holds for $G_m^+$.

Under the Bayesian model, $P^\star$ is the realized value
of the random kernel $P$.
Using $\|\mathbf x\|_1^2\le S\|\mathbf x\|_2^2$
and the posterior variance identity gives
\begin{align*}
  \mathbb E\!\left[
    (\Delta_m^P)^2\mid\mathcal F_{t_m}
  \right]
  &\le
  S\sum_{s,a}
  \mathbb E\!\left[
    \|P(\cdot\mid s,a)
      -\mu_{t_m}(\cdot\mid s,a)\|_2^2
    \,\middle|\,\mathcal F_{t_m}
  \right]\\
  &=S\sum_{s,a}V_{t_m}(s,a).
\end{align*}
Conditional Markov's inequality therefore yields an event
$\mathcal E_m$ with posterior probability at least
$1-\delta$ on which
\[
  \Delta_m^P
  \le
  \left(
    \frac{S}{\delta}\sum_{s,a}V_{t_m}(s,a)
  \right)^{1/2}.
\]
Combining the two bounds on
$\mathcal E_m\cap\mathcal H_m$ gives $G_m^+\le R_m$.
Finally, using
\[
  V_{t_m}(s,a)
  =
  \frac{c_{t_m}(s,a)}{\beta_{t_m}(s,a)+1},
  \qquad
  c_{t_m}(s,a)\le1,
\]
and
$\beta_{t_m}(s,a)\ge\beta_{\min}+\nu t_m$
proves \eqref{eq:mean-gap-rate}.
\end{proof}

\section{Bounding $\epsilon_m$}
\label{appendix:rest_epsilon_bound}
Adding and subtracting $\mathbf w_m$ in the oracle error
gives, on the same event,
\begin{equation}
\begin{aligned}
  \epsilon_m
  &=
  G_m+
  \left\langle
    \nabla U_\kappa(\widehat{\mathbf d}_{t_m})
    -\mathbf w_m,\,
    \boldsymbol{\psi}_m^\star-\boldsymbol{\psi}_m
  \right\rangle\\
  &\le
  R_m+
  2\left\|
    \nabla U_\kappa(\widehat{\mathbf d}_{t_m})
    -\mathbf w_m
  \right\|_\infty,
\end{aligned}
\end{equation}
where we used
$\|\boldsymbol{\psi}_m^\star-\boldsymbol{\psi}_m\|_1\le2$.

Under the sampling condition,
$\widehat d_{t_m}(s,a)\ge\nu$, and
\[
  t_mV_{t_m}(s,a)
  =
  \frac{c_{t_m}(s,a)}
       {\widehat d_{t_m}(s,a)+(\beta_0(s,a)+1)/t_m}
  \le\nu^{-1}.
\]
Thus, one may take $W=\nu^{-\kappa}$.

To control the gradient error, first observe that
\[
  |c_{t_m}(s,a)-c^\star(s,a)|
  \le
  2\|\mu_{t_m}(\cdot\mid s,a)
       -P^\star(\cdot\mid s,a)\|_1
  \le2\Delta_m^P.
\]
Let $b_{\max}:=\max_{s,a}(\beta_0(s,a)+1)$.
Since $c^\star(s,a),c_{t_m}(s,a)\in[0,1]$, we obtain
\[
  \left|
    \frac{c^\star(s,a)}{\widehat d_{t_m}(s,a)}
    -
    \frac{c_{t_m}(s,a)}
         {\widehat d_{t_m}(s,a)+(\beta_0(s,a)+1)/t_m}
  \right|
  \le
  \frac{2\Delta_m^P}{\nu}
  +\frac{b_{\max}}{\nu^2t_m}.
\]
The map $x\mapsto x^\kappa$ is
$\kappa\nu^{-(\kappa-1)}$-Lipschitz on $[0,\nu^{-1}]$.
Therefore,
\begin{equation}
  \left\|
    \nabla U_\kappa(\widehat{\mathbf d}_{t_m})
    -\mathbf w_m
  \right\|_\infty
  \le
  2\kappa\nu^{-\kappa}\Delta_m^P
  +
  \frac{\kappa b_{\max}}{\nu^{\kappa+1}t_m}.
\end{equation}
Combining this bound with the planning-discrepancy bound
yields, on the same event,
\begin{equation}
  \epsilon_m
  \le{}
  (C_{\rm occ}+4\kappa)\nu^{-\kappa}S
  \sqrt{
    \frac{A}
         {\delta(\beta_{\min}+\nu t_m+1)}
  }
  +
  \frac{2\kappa b_{\max}}{\nu^{\kappa+1}t_m}.
\label{eq:oracle-error-rate}
\end{equation}
Hence, the positive part
$\epsilon_m^+:=\max\{\epsilon_m,0\}$
is $O(t_m^{-1/2})$ at each fixed episode and fixed
posterior confidence, with all other parameters held
constant.

\section{Proof of Lemma~\ref{lem:posterior-coefficient-error}}
\label{appendix:proof_lem:posterior-coefficient-error}

\begin{proof}
Under the fixed-kernel law,
each transition row can be sampled from its own sequence of
independent draws from $P^\star(\cdot\mid s,a)$. Write
$\widehat p_n(\cdot\mid s,a)$ for the empirical distribution of
the first $n$ draws for that row. Coordinatewise Hoeffding bounds
and a union bound over the $S^2A$ coordinates and $1\le n\le T$
give an event of probability at least $1-\delta/2$ on which
\[
  \bigl\|\widehat p_n(\cdot\mid s,a)-P^\star(\cdot\mid s,a)\bigr\|_2
  \le \sqrt{\frac{S q_T(\delta)}{2n}}
\]
simultaneously for all rows and counts. The bound therefore applies
at the adaptive count $n=N_T(s,a)$.

By \eqref{eq:beta-mu}, the posterior mean mixes $\widehat p_n$ and
$\mu_0$ with prior weight $\beta_0(s,a)/(n+\beta_0(s,a))$.
Two probability vectors are at Euclidean distance at most $\sqrt{2}$.
Also, the difference in their squared norms is at most twice their
Euclidean distance. Consequently, on the same event and for
$n=N_T(s,a)>0$,
\[
  |c_T(s,a)-c^\star(s,a)|
  \le \sqrt{\frac{2S q_T(\delta)}{n}}
       +\frac{2\sqrt{2}\,\beta_0(s,a)}{n}.
\]
When $N_T(s,a)\ge\eta T$ for all $(s,a)$, dividing this inequality
by $N_T(s,a)+\beta_0(s,a)+1$ and taking the maximum gives
\[
  \max_{s,a}\frac{|c_T(s,a)-c^\star(s,a)|}
                    {N_T(s,a)+\beta_0(s,a)+1}
  \le \frac{1}{\eta T+\min_{s,a}\beta_0(s,a)+1}
  \left(
    \sqrt{\frac{2S q_T(\delta)}{\eta T}}
    +\frac{2\sqrt{2}\,\beta_{\max}}{\eta T}
  \right).
\]
\end{proof}

\section{Proof of Theorem~\ref{thm:bayesian-kappa-main}}
\label{appendix:proof_thm_bayesian-kappa-main}
Use the joint event from Section~\ref{subsec:optimization} on which
$\widehat d_T(s,a)\ge\eta$ for all $(s,a)$ and
$U_\kappa(d_\kappa^\star)-U_\kappa(\widehat d_T)
\le\varepsilon_{\mathrm{opt}}(T,\delta)$. Given its probability
bound of $1-\delta/2$, its intersection with the event in
Lemma~\ref{lem:posterior-coefficient-error} has probability at
least $1-\delta$.

Define $F(\mathbf d) = \max_{s, a}\frac{c^\star(s, a)}{d(s, a)}$.
On this intersection, Lemma~\ref{lem:posterior-coefficient-error} give
$M_T\le F(\widehat d_T)/T+\varepsilon_{\mathrm{post}}(T,\delta)$.
For $\kappa>1$, Proposition~\ref{prop:kappa-approximation} then
yields the sum of the utility bound and the posterior error,
\begin{equation}
  M_T
  \le \frac{\eta^{-1/\kappa}}{T}
       \bigl[(1-\kappa)U_\kappa(\widehat d_T)\bigr]^{1/\kappa}
      +\varepsilon_{\mathrm{post}}(T,\delta).
  \label{eq:variance-utility-posterior-bound}
\end{equation}
Multiplying the convergence gap inequality by $\kappa-1$ and
rearranging, we obtain
\begin{align*}
  (1-\kappa)U_\kappa(\widehat d_T)
  &\le (1-\kappa)U_\kappa(d_\kappa^\star)
       +(\kappa-1)\varepsilon_{\mathrm{opt}}(T,\delta)\\
  &\le (1-\kappa)U_\kappa(\mathbf d^\star)
       +(\kappa-1)\varepsilon_{\mathrm{opt}}(T,\delta)\\
  &\le (C^\star)^\kappa
       +(\kappa-1)\varepsilon_{\mathrm{opt}}(T,\delta).
\end{align*}
The second inequality uses the optimality of $d_\kappa^\star$ for
$U_\kappa$ and $1-\kappa<0$. The third uses the upper bound in
Proposition~\ref{prop:kappa-approximation} at $\mathbf d^\star$.
Substituting this into \eqref{eq:variance-utility-posterior-bound}
gives
\[
  M_T
  \le \frac{\eta^{-1/\kappa}}{T}
      \left[(C^\star)^\kappa+
           (\kappa-1)\varepsilon_{\mathrm{opt}}(T,\delta)\right]^{1/\kappa}
      +\varepsilon_{\mathrm{post}}(T,\delta),
\]
which is \eqref{eq:bayesian-kappa-main-bound}. For $\kappa=1$, the
visit floor gives $\eta F(\widehat d_T)\le\sum_{s,a}c^\star(s,a)$,
and $\sum_{s,a}c^\star(s,a)\le F(\mathbf d^\star)=C^\star$
by averaging the ratios with weights $\mathbf d^\star$. Hence
\eqref{eq:bayesian-kappa-main-bound} also holds at $\kappa=1$.

When $C^\star>0$, concavity of $x^{1/\kappa}$ gives, for $b\ge0$,
\[
  \left[(C^\star)^\kappa+b\right]^{1/\kappa}
  \le C^\star+\frac{b}{\kappa(C^\star)^{\kappa-1}}.
\]
Taking $b=(\kappa-1)\varepsilon_{\mathrm{opt}}(T,\delta)$ yields
\begin{align*}
  M_T
  &\le \frac{\eta^{-1/\kappa}C^\star}{T}
      +\frac{\eta^{-1/\kappa}(\kappa-1)\varepsilon_{\mathrm{opt}}(T,\delta)}
             {\kappa T(C^\star)^{\kappa-1}}
      +\varepsilon_{\mathrm{post}}(T,\delta)\\
  &= \frac{C^\star}{T}
     +\frac{(\eta^{-1/\kappa}-1)C^\star}{T}\\
  &\quad
     +\frac{\eta^{-1/\kappa}(\kappa-1)\varepsilon_{\mathrm{opt}}(T,\delta)}
            {\kappa T(C^\star)^{\kappa-1}}
     +\varepsilon_{\mathrm{post}}(T,\delta),
\end{align*}
which gives the three error terms in \eqref{eq:bayesian-kappa-expanded-bound}.

For the rate form, write the convergence estimate from
Section~\ref{subsec:optimization} as
$\varepsilon_{\mathrm{opt}}(T,\delta)\le K_\delta T^{-1/3}$,
where $K_\delta$ is independent of $T$. Bounding the denominator in
\eqref{eq:bayesian-kappa-posterior-error} below by $\eta T$ gives
\[
  \varepsilon_{\mathrm{post}}(T,\delta)
  \le \sqrt{\frac{2S q_T(\delta)}{\eta^3}}\,T^{-3/2}
      +\frac{2\sqrt{2}\,\beta_{\max}}{\eta^2}T^{-2}.
\]
Inserting these estimates in \eqref{eq:bayesian-kappa-expanded-bound}
gives, for $C^\star>0$,
\begin{align*}
  M_T
  &\le \frac{C^\star}{T}
      +\frac{(\eta^{-1/\kappa}-1)C^\star}{T}\\
  &\quad
      +\frac{\eta^{-1/\kappa}(\kappa-1)K_\delta}
             {\kappa(C^\star)^{\kappa-1}}T^{-4/3}
      +\sqrt{\frac{2S q_T(\delta)}{\eta^3}}\,T^{-3/2}
      +\frac{2\sqrt{2}\,\beta_{\max}}{\eta^2}T^{-2}.
\end{align*}
\end{document}